\def\year{2020}\relax
\documentclass[letterpaper]{article} 
\usepackage{aaai20}  
\usepackage{times}  
\usepackage{helvet} 
\usepackage{courier}  
\usepackage[hyphens]{url}  
\usepackage{graphicx} 
\usepackage{graphicx}  
\usepackage{amsmath}
\usepackage{color}
\usepackage{amssymb}
\usepackage{array,multirow}
\newtheorem{theorem}{Theorem}
\newcommand{\comment}[1]{}
\usepackage{subfigure}
\newcommand{\etal}{\textit{et al. }}

\title{SubSpace Capsule Network }

\author{Marzieh Edraki,\textsuperscript{\rm 1}\thanks{Corresponding author} \Large \textbf{Nazanin Rahnavard,\textsuperscript{\rm 1,2} Mubarak Shah\textsuperscript{\rm 1}}\\ 
\textsuperscript{\rm 1} Center for Research in Computer Vision\\
\textsuperscript{\rm 2}Department of Electrical and Computer Engineering\\University of Central Florida\\ Orlando, Florida, USA, 32816\\ 
{m.edraki@knights.ucf.edu, nazanin@eecs.ucf.edu, shah@crcv.ucf.edu} 
}
 
\begin{document}

\maketitle
\begin{abstract}
 Convolutional neural networks (CNNs) have become a key asset to most of fields in AI. Despite their successful performance, CNNs suffer from a major drawback. They fail to capture the hierarchy of spatial relation among different parts of an entity.
 As a remedy to this problem, the idea of capsules was proposed by Hinton.
In this paper, we propose the SubSpace Capsule Network (\textbf{SCN}) that exploits the idea of capsule networks to model possible variations in the appearance or implicitly-defined properties of an entity through a group of capsule subspaces instead of simply grouping neurons to create capsules. A  capsule is created by projecting an input feature vector from a lower layer onto the capsule subspace using a learnable transformation. This transformation finds the degree of alignment of the input with the properties modeled by the capsule subspace.
 

We show that \textbf{SCN} is a general capsule network that can successfully be applied to both discriminative and generative models without incurring computational overhead compared to CNN during test time. 
Effectiveness of \textbf{SCN} is evaluated through a comprehensive set of experiments on supervised image classification, semi-supervised image classification  and  high-resolution image generation tasks using the generative adversarial network (GAN) framework. \textbf{SCN} significantly improves the performance of the baseline models in all 3 tasks.
\end{abstract}
\section{Introduction}
In the recent years, convolutional neural networks (CNNs) have become a key asset to most of fields in AI. Various tasks in computer vision, reinforcement learning, natural language and speech processing systems have achieved significant improvement by using them. New applications like music generation~\cite{dong2018musegan}, visual text correction \cite{mazaheri2018visual}, online fashion recommendation \cite{han2017learning} are founded on the feature learning capability of CNN architectures.  
Despite their successful performance, CNNs suffer from a major drawback.  
They fail to capture the hierarchy of spatial relation among different parts of an entity. As a remedy to this problem, Hinton \etal introduced the idea of  \textit{Capsule Networks}~\cite{hinton2011transforming}. Capsule networks received a flurry of attention after achieving the state-of-the-art performance on image classification \cite{sabour2017dynamic}, text classification \cite{zhao2018investigating}, action detection and localization~\cite{duarte2018videocapsulenet}, image segmentation tasks \cite{lalonde2018capsules}, etc. Moreover, many efforts have been made to improve the structure of capsule networks  \cite{hinton2018matrix} \cite{bahadori2018spectral}\cite{zhang2018cappronet} as a new generation of deep neural networks.

A capsule is defined as a group of neurons that can ultimately model different properties such as pose, texture or deformation of an entity or a part of an entity. Each layer of a capsule network consists of many capsules. In a well-trained capsule network, activation vector of each capsule represents the instantiation parameters of the entity and the length of the capsule scores the presence of that feature or part of that entity
In this paper, while we still follow the main definition of capsules, we propose \textit{Subspace Capsule Networks} ({\bf SCNs}), which build capsules based on the degree of relevance of an input feature vector to {\em a group of learned subspaces} .
In \emph{SCN}s, corresponding to each entity or part of that entity, a specific capsule subspace is learned. 
Then a capsule is created by projecting the input feature vector onto the capsule subspace using a learned transformation, defined based on the basis of the corresponding capsule subspace. 
Intuitively speaking, a capsule subspace captures the variation
in visual properties; like appearance, pose, texture and deformation; 
of an object or an implicitly defined feature of that object 
 The length of the output vector of a subspace capsule represents the degree of alignment of an input with the properties modeled by that subspace. Hence, if a subspace capsule has a large activity vector, it means that the input feature vector is highly related to the entity modeled by that subspace and vice versa. This form of creating subspace capsules makes it independent of any form of routing required in capsule network introduced in \cite{sabour2017dynamic} or \cite{hinton2018matrix}. Due to this property,  \emph{SCN} is easily scalable to large network architectures and large datasets. 

Closest to our work is the CapProNet model, proposed by \cite{zhang2018cappronet}, in which authors
apply subspace-based capsules \emph{merely} in the last layer of an image classification network and only require capsule length for prediction. In the classification task with $N$ classes, a group of capsule subspaces $\{\mathcal{S}_1,...\mathcal{S}_N\}$ are learned. Then the capsule corresponding to each class is created by orthogonal projection of the input feature vector from backbone model onto the learned subspace. The input image belongs to the class with the largest capsule length. 
In \emph{SCN}, unlike CapProNet we are interested in both subspace capsules and norm of capsules. 

The summary of our contributions is as follows: 
\begin{quote}
\begin{itemize}
\item The proposed  \emph{SCN} is a general capsule model that can be used without any change in the formulation in both \emph{generative models} as well as \emph{discriminative  models}.
\item \emph{SCN} is computationally efficient with no computational overhead during test phase and a negligible computational overhead with help of the method introduced in Section (\ref{Implementation_det}) during training, compared to the baselines.
\item When applied in generator model of a GAN, \emph{SCN} consistently improves the relative FID score of generated samples by at least $20\% $ in all of our experiments.
\item \emph{SCN} achieves state-of-the-art performance in semi-supervised classification of CIFAR10 and SVH datasets and improves the relative error rate of the baseline models by at least $23\%$ for these 2 datasets.  
\item \emph{SCN} is easily scalable to large architectures and datasets like ImageNet. When applies on the last block of the Resnet model, it decreases the Top-1 error rate by $5\%$ relatively.
\end{itemize}
\end{quote}
The rest of the paper is organized as follows. We first briefly review some of the related studies with capsule networks and GAN models in Section \ref{Related Works}. Subspace Capsule Network is formally presented in Section~\ref{SCN} followed by studying the effects of \emph{SCN} on the GAN framework in Section~\ref{Application}. Implementation details are discussed in Section ~\ref{Implementation_det}. We evaluate the performance of \emph{SCN} in Section ~\ref{Exper} and the conclusion is presented in Section \ref{conclud}.

\section{Related Work} \label{Related Works}
The idea of capsule networks was first introduced in Transforming auto-encoders, where Hinton \etal pointed out that CNNs cannot achieve viewpoint invariance just by looking at the activity of a single neuron, and a more complex structure like a capsule is necessary. Output of a capsule is a vector that summarizes information about an entity or part of that entity. The main advantages of capsule networks is that the part-whole relation can be captured through the capsules of consecutive layers.
\cite{sabour2017dynamic} define a capsule as a group of neurons, whose orientation of its output vector represents the instantiation parameters of a visual entity modeled by that capsule and its length represents the probability of entity's existence. They use dynamic routing between capsules to capture the part-whole relationship. Dynamic routing works based on measuring the agreement of two capsule in consecutive layers using scalar product of their capsule vectors.
The subsequent paper~\cite{hinton2018matrix} extends the idea of capsule by separating it into a $4\times 4$ pose matrix and an activation probability. Dynamic routing is updated to EM-routing algorithm, which is a more efficient way in measuring the agreement between capsules. The new capsule structure leads to the state-of-the-art performance in classification task of SmallNorb dataset. \cite{singh2019dual} use capsule idea in low-resolution image recognition.
\begin{figure}
\centering
  \includegraphics[width=0.75\linewidth]{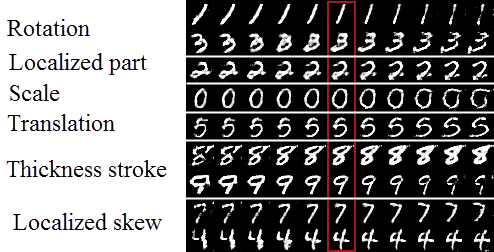}
\caption{\small{MNIST samples are generated by varying each dimension of capsules of first layer by a value in the range of $[-2.5,2.5]$}}
\label{fig:mnist_tweak}
\end{figure}
The idea of 3D capsules introduced in \cite{duarte2018videocapsulenet} to tackle action detection and localization problem. CapProNet~\cite{zhang2018cappronet} proposes learning a group of capsule subspaces in the final layer of a CNN for the image classification task.
Capsule network was also applied on medical image segmentation task by \cite{lalonde2018capsules} and achieve competitive results based on the convolution-deconvolution capsule network structure. 

The common point among all of these studies is that they all try to solve a discriminative task, like classification, image segmentation, and action detection using a capsule network. There have been a few attempts in using the capsule architecture in a generative model. CapsuleGAN \cite{jaiswal2018capsulegan} applies capsule network in the discriminator of a GAN to improve the quality of generated samples and CapsPix2Pix ~\cite{bass2019image} uses convolution capsules to synthesise images conditioned on segmentation labels to pre-train segmentation models for the medical image segmentation task.  
Since the introduction of GANs by \cite{goodfellow2014generative}, many efforts have been made to improve the stability of training and quality of generated samples. Among them Wasserstein loss with gradient penalty \cite{arjovsky2017wasserstein}\cite{gulrajani2017improved} and Spectral Normalization \cite{miyato2018spectral} successfully stabilize
the training process by enforcing Lipschitz continuity on discriminator and ProgressiveGAN \cite{karras2017progressive} and BigGAN \cite{brock2018large} generate high-quality samples by improving the architecture. 

In this paper, we exploit the inherent property of subspace capsules, which is modeling the variation in appearance of a visual entity in a GAN model, to produce diverse and high-quality image samples. We also show the superiority of our proposed model in semi-supervised image classification using the GAN framework and also supervised image classification. 
\section{Subspace Capsule Networks}\label{SCN}
In this section, we formalize the idea of subspace capsule networks (\textbf{SCNs}) by presenting their main components. In each layer, a \emph{SCN} learns a group of capsule subspaces, each of which captures possible variations of an implicitly defined visual entity. An input from the lower layer is projected onto each of these capsule subspaces to create new capsules. If the input and a capsule subspace are related; for instance, the input is a part of the entity represented by a capsule subspace; the output vector (projection of the input vector on to the the corresponding subspace) will be large. Moreover, the orientation of a capsule vector represents the properties of that entity.

Since the key component of a \emph{SCN} is finding the level of alignment of input feature vectors and capsule subspaces, we elaborate on the proposed projection matrix and formulate subspace capsules. Then, capsule activation functions are presented followed by subspace capsule convolution layer and the idea of subspace capsule mean pooling.
 \subsection{Projection onto a Capsule Subspace}
 For the layer $k$,
 suppose $\boldsymbol{x}\in\mathbb{R}^{d}$ is an input feature vector from a lower layer $k-1$. Suppose a capsule subspace $\mathcal{S}$ with dimensions $c$ is formed as the span of the columns of the weight matrix $\boldsymbol{W}\in\mathbb{R}^{d\times c}$, where $c\ll d$.
 
 The most straight-forward way to find the degree of alignment of feature vector $\boldsymbol{x}$ and capsule subspace $\mathcal{S}$ is to orthogonally project $\boldsymbol{x}$ onto subspace $\mathcal{S}$.
 This problem has a closed-form solution as follows
 \begin{equation}\label{Eq:projection}
\boldsymbol{y}=\underbrace{\boldsymbol{W}(\boldsymbol{W}^T\boldsymbol{W})^{-1}\boldsymbol{W}^T}_{P}\boldsymbol{x},\\
\end{equation}
where $\boldsymbol{P}\in \mathbb{R}^{d\times d}$ is the matrix of orthogonal projection onto $\mathcal{S}$, and $\boldsymbol{y}\in \mathbb{R}^{d}$ is the projection of $\boldsymbol{x}$ onto $\mathcal{S}$. The larger the length of $\boldsymbol{y}$, the more correlated $\boldsymbol{x}$ and capsule subspace $\mathcal{S}$ are. In other words, $\boldsymbol{x}$ has more of the properties modeled by $\mathcal{S}$.
 
 However, the projection matrix $\boldsymbol{P}\in \mathbb{R}^{d\times d}$ has the major drawback of being a square matrix. This means that if we create a capsule by projecting the feature vector $\boldsymbol{x}\in \mathbb{R}^{d}$ onto capsule subspace $\mathcal{S}$ using $\boldsymbol{P}$, that capsule is still in the $d$-dimensional space. Practically speaking, if $d$ is large, which is usually the case in deep models, having different capsule types using the orthogonal projection matrix $\boldsymbol{P}$ would be impossible since it demands a lots of memory.     
 To be able to benefit from various sizes of capsules through the sequence of subspace capsule layers, one needs a transformation that allows the input feature vector $\boldsymbol{x}$ to be mapped onto the $c$-dimensional space of capsule subspace, while it still preserves the relation among the capsules in the consecutive layers of network.  
 We propose to employ an \emph{intermediate domain} indicated by a \emph{transformation matrix} $\boldsymbol{P}_c\in\mathbb{R}^{c \times d}$ in order to  exploit capsule subspaces. This matrix is derived by decomposing the orthogonal projection matrix $\boldsymbol{P}$ as 
  \begin{equation}\label{Eq:decompse}
\boldsymbol{P}=\boldsymbol{P}_d~\boldsymbol{P}_c,\\
\end{equation}
where
\begin{subequations}
\begin{align}\label{Eq:projection_d}
\boldsymbol{P}_d&=\boldsymbol{W}(\boldsymbol{W}^T\boldsymbol{W})^{-1/2},\\
\boldsymbol{P}_c&=(\boldsymbol{W}^T\boldsymbol{W})^{-1/2}\boldsymbol{W}^T. \label{Eq:projection_c}
\end{align}
\end{subequations}
Here $\boldsymbol{P}_c$ is the transformation that maps the input feature vector $\boldsymbol{x}$ into the $c$-dimensional capsule space\footnote{This $c$-dim space is defined by the span of right singular vectors of $\boldsymbol{W}$.}, and $\boldsymbol{P}_d$ is the transformation that projects vectors in the capsule space back to the original $d$-dimensional space of input feature vector $\boldsymbol{x}$. 
Now, the capsule that corresponds to the capsule subspace $\mathcal{S}$ can be created by projecting feature vector $\boldsymbol{x}$ onto the $c$-dimensional capsule space as
  \begin{equation}\label{Eq:capsule}
\boldsymbol{u}=\boldsymbol{P}_c~\boldsymbol{x}. 
\end{equation}
Here, $\boldsymbol{u}$ indicates the low-dimensional representation of $\boldsymbol{x}$ in the capsule space.  Matrix $\boldsymbol{P}$ is a semi-definite and symmetric matrix. Thus its decomposition as suggested in Equation (\ref{Eq:decompse}) has special properties. We \textit{claim} that a capsule created using $\boldsymbol{P}_c$ has the same information about the instantiation parameters and also the score of presence of features, as it would be created by transformation $\boldsymbol{P}$.
Proof of this claim follows from the Theorem~\ref{thm1}.  
\begin{theorem}\label{thm1}
Let $P$, as defined in~(\ref{Eq:projection}), denote an orthogonal projection matrix onto the subspace spanned by columns of the weight matrix $\boldsymbol{W}\in\mathbb{R}^{d \times c}$. Assume $P$ is decomposed into two matrices $\boldsymbol{P}_d$ and $\boldsymbol{P}_c$ as in~(\ref{Eq:decompse}). Then, the transformation matrix $\boldsymbol{P}_d$ is an \textit{isomorphic} transformation between $\mathbb{R}^c$ and $\mathbb{R}^d$, i.e., $\forall \boldsymbol{u}\in \mathbb{R}^c, \|\boldsymbol{u}\|_2=\|\boldsymbol{P}_d\boldsymbol{u}\|_2$.

\end{theorem}
The following can be concluded from Theorem~\ref{thm1}.
\begin{itemize}
    \item  The norm of the capsule vector $\boldsymbol{u}$ defined in Equation (\ref{Eq:capsule}) represents the score of 
    features modeled by $\mathcal{S}$ in the input feature vector $\boldsymbol{x}$, since  $\|\boldsymbol{u}\|=\|\boldsymbol{y}\|$, where $||.||$ denotes that $l_2$-norm of a vector.
    \item For two input feature vectors $\boldsymbol{x}_1$ and $\boldsymbol{x}_2$, the relation between their corresponding capsules $\boldsymbol{u}_1$ and $\boldsymbol{u}_2$ is the same as the relation between $\boldsymbol{y}_1$ and $\boldsymbol{y}_2$. For instance the angle between $\boldsymbol{u}_1$ and $\boldsymbol{u}_2$ is the same as the angle between $\boldsymbol{y}_1$ and $\boldsymbol{y}_2$. 
\end{itemize}

\begin{figure*}
\begin{center}
\subfigure[Generator]{
\begin{minipage}{.75\textwidth}
  \centering
  \includegraphics[width=0.8\linewidth]{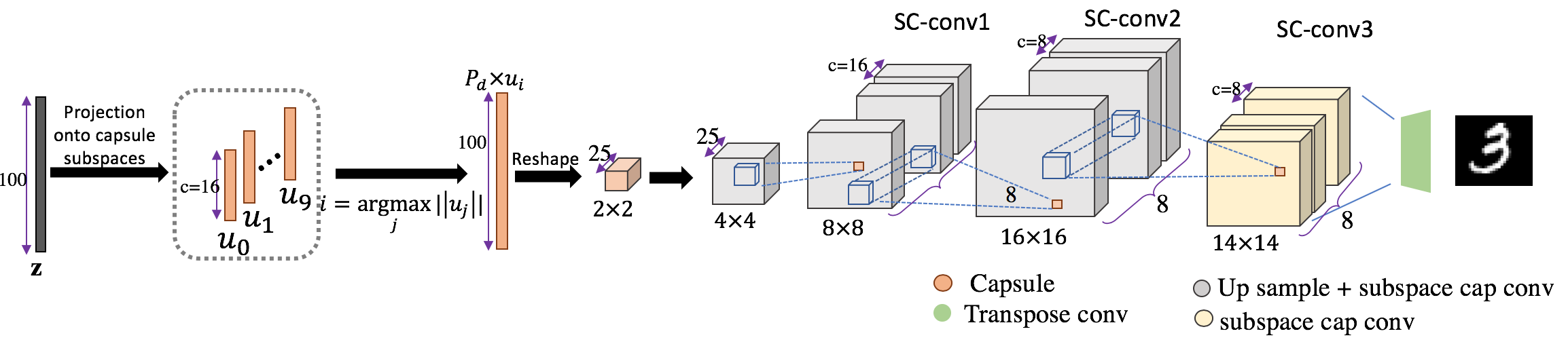}
  \label{fig:MNIST_Gen}
\end{minipage}}

\subfigure[Discriminator]{
\begin{minipage}{.75\textwidth}
  \centering
  \includegraphics[width=0.8\linewidth]{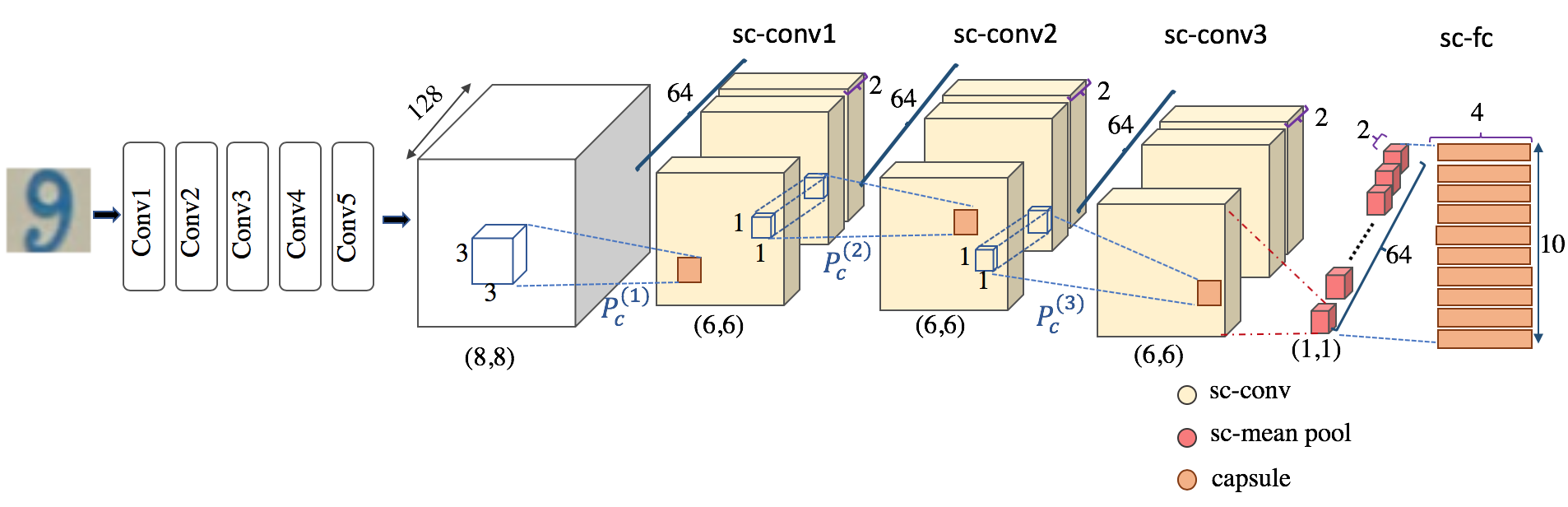}
  \label{fig:SVHN_disc}
\end{minipage}}
 \end{center}
\caption{\small{a) In the generator,
   the latent representation $\boldsymbol{z}$ is projected onto 10 capsule sbspaces with dimension $c=16$ in the first layer. The capsule with largest vector is selected and reshaped to a cube of $25\times 2 \times 2$, then up-sampled to double the spatial resolution to $ 4 \times 4$. This cube goes through 2 layers of sc-conv with 8 capsule types of  16 and 8 capsule dimensions, respectively, each followed by upsampling operation to get to the resolution of $16\times 16$. The final sc-conv layer has 8 subspace capsule types each with 8 dimension. The output of this layer is fed to a transposed convolution layer to generate the final image. b) The SCN architecture of  discriminator component of GAN for SVHN dataset. Features are extracted using 6 convolutional layers, followed by 3 subspace capsule convolution (SC-conv) layers each with 64 subspace capsule types, one subspace capsule mean pool(SC-mean pool) layer and final subspace capsule fully connected (SC-Fc) layer with 10 capsule types.}}
   \vspace{-3mm}
\label{fig:celeba_bedroom_horse}
\end{figure*}

\subsection{Activation Function}
We apply two types of activation functions on subspace capsule based on our interpretation of the length of the output vector of a capsule.
The length of the output vector of a capsule can be interpreted from the \emph{confidence} perspective. A high confidence for a capsule shows that the input feature vector is highly aligned with the capsule subspace. In other words, the input feature vector contains the entity that is modeled by capsule subspace. We also want to suppress the effect of noisy capsules of layer $L$ on activating the capsules of the next layer. 
Following this perspective we propose ``\textbf{sparking}" function given by
\begin{equation}
\boldsymbol{v}=max(\|\boldsymbol{u}\|-b^2, 0)\frac{\boldsymbol{u}}{\|\boldsymbol{u}\|},
\end{equation}
where $b$ is a parameter that can be learned.

Intuitively, the proposed activation function tries to increase the capsule certainty, if $\boldsymbol{x}$ is related to the entity modeled by capsule subspace $\mathcal{S}$, or completely turn the capsule off if the length of it is below the threshold $b^2$. We initialize $b^2=0.25$ in our experiments and update it along with network parameters through the training process using stochastic gradient decent method. 

Another possibility is to relate the \emph{probability} of the presence of
an entity modeled by a capsule subspace by the length of the output capsule. For that, we follow~\cite{sabour2017dynamic} and use squashing function defined as  
\begin{equation}
\boldsymbol{v}=\frac{\|\boldsymbol{u}\|^2}{1+\|\boldsymbol{u}\|^2}\frac{\boldsymbol{u}}{\|\boldsymbol{u}\|}.
\end{equation}

\comment{, where we want the length of the output vector of a capsule to represent the \emph{probability} of the presence of
the entity modeled by the capsule subspace. Then we follow \cite{sabour2017dynamic} and use squashing function defined as  
\begin{equation}
\boldsymbol{v}=\frac{\|\boldsymbol{u}\|^2}{1+\|\boldsymbol{u}\|^2}\frac{\boldsymbol{u}}{\|\boldsymbol{u}\|}.
\end{equation}}

We found \emph{sparking} function is more effective in discriminative tasks, i.e., in our (semi-)supervised classification of images; since it outputs sparse feature maps by turning off noisy capsules which leads to faster convergence. Noisy capsules in each layer are those capsules represent the properties that are not related to the input image and would have a small activity vector. 
While in generative models, having small but non-zero values by applying squashing activation function on capsules leads to the higher quality of generated samples.

\subsection{Subspace Capsule Convolution}
\emph{SCN} can also benefit from the idea of weight sharing of CNNs by using the same subspace capsule types in all spatial locations of an image. 

In subspace capsule convolution, if the input $\boldsymbol{x}$ has $i$ feature maps, and we want to create a $c$ dimensional subspace capsule convolution kernel with receptive field of $k$, we need to build the transformation matrix $\boldsymbol{P}_c$ as defined in Equation~(\ref{Eq:projection_c}), based on a weight matrix $\boldsymbol{W}\in \mathbb{R}^{(i\times k\times k)\times c}$.  We can treat each row of the projection matrix $\boldsymbol{P}_c$ as one convolution kernel of size $(i\times k \times k)$, that convolves over input feature maps and generates a single element of output capsule.
So if $\boldsymbol{P}_c$ gets reorganized into a 4-dimensional tensor with the shape of $(c\times i\times k\times k) $, then it can be used as the kernel of regular convolution operation and the capsule corresponds to each spatial location would be placed along the output feature maps. 
Now, if we want to have $n$ subspace capsule types, we can create a group of projection matrices $\{\boldsymbol{P}_{c_1},...,\boldsymbol{P}_{c_n}\}$, after reorganizing each of them to a 4-dimensional tensor, and then  concatenate them to create a kernel of shape $(nc\times i\times k\times k)$. From now on, we represent the kernel of a Subspace capsule convolution layer with a tuple of $(n,c,k,k)$.

\subsection{Subspace Capsule Mean Pooling}
\noindent The idea of mean pooling comes naturally after subspace capsule convolutions. In subspace capsule convolution, capsules of the same type represent the same visual property regardless of spatial positions. So it is a safe assumption that capsules of the same type in a small receptive field of $k\times k$ have similar orientation and a single capsule with mean of those capsule vectors can represent all of them.  
\subsection{Gradient Analysis}
Subspace capsule networks are trained based on the stochastic gradient descent methods. So analyzing the gradient that is used to update $\boldsymbol{W}$ in each step clarifies how \emph{SCN} learns capsule subspaces.

Assume we have a loss function $L$ and we want to differentiate it with respect to the weight matrix $\boldsymbol{W}$, the basis of subspace $\mathcal{S}$, through the projection onto subspace~(Equation  \ref{Eq:projection_c}).
For the sake of simplicity we first assume a $1-$dimensional capsule subspace, i.e, $c=1$. Using the chain rule the gradient is given by: 
\begin{gather}
    \nabla_W L= \frac{1}{\|\boldsymbol{W}\|} (\boldsymbol{I}-\boldsymbol{P})~ \nabla_{P_c} L,
\end{gather}
\noindent where $\nabla_{P_c}L$ is the gradient with respect to the projection matrix $\boldsymbol{P}_c$ and it is computed the same way as the gradient with respect to the kernel of a convolution operation. 
The term $(\boldsymbol{I}-\boldsymbol{P})$ is the projection matrix onto the orthogonal complement of subspace $\mathcal{S}$. This shows that the basis of capsule subspace $\mathcal{S}$ spanned by the columns of $\boldsymbol{W}$ only updated along the orthogonal complement of $\mathcal{S}$ up to the scale 
$\frac{1}{\|\boldsymbol{W}\|}$.
The orthogonal complement of $\mathcal{S}$ can contain those novel features from $\boldsymbol{x}$ that are not yet captured by $\mathcal{S}$. 

This nice property of gradient can extend to higher dimensional subspaces. 
Using the chain rule and derivative of inverse of a matrix~(Petersen \etal ) the gradient is as follows:
\begin{gather}
    \nabla_{{W}_{ij}} L=(\boldsymbol{W}^T \boldsymbol{W})^{-\frac{1}{2}} \boldsymbol{s}^{T}_{ij} (\boldsymbol{I}-\boldsymbol{P})~ \nabla_{P_c} L,
\end{gather}
where $\boldsymbol{s}_{ij}$ is a single non-zero entry matrix corresponding to the gradient of $\boldsymbol{W}$ with respect to one of its elements in position $(i,j)$. The general case also supports our conclusion from the special case since $(\boldsymbol{W}^T \boldsymbol{W})^{\frac{-1}{2}}$ only stretch the space along the basis of subspace by the scale factor of eigenvalues of $(\boldsymbol{W}^T \boldsymbol{W})^{-\frac{1}{2}}$.

\section {SubSpace Capsule Networks for GANs} \label{Application}
So far we have defined all the building blocks of a subspace capsule network. Next, we want to discuss how \emph{SCN} can be effective in enhancing the performance of GANs. When GAN models are used in semi-supervised learning tasks, like image classification, the \emph{discriminator} can benefit from \emph{SCN} ability by modeling the possible variations of visual properties; for instance texture, pose, color corresponding to an entity using a group of capsule subspaces through a sequence of subspace capsule layers. By creating the capsule using projection of input feature vector onto these capsule subspaces, and considering the length of capsules as confidence about the presence of those properties that are modeled by subspaces, the discriminator can be made invariant with respect to the possible deformations of each visual property.
GAN models can also leverage the ability of subspace capsule layer in the \emph{generator} network. A subspace capsule generator consists of multiple subspace capsule layers and each layer has multiple subspace capsule types. When trained, each subspace capsule type models all the possible variation of a visual entity. Now the goal of the generator in each layer is to find the related properties and features that need to be added to the generated image so far.  
In addition using \emph{SCN} as generator leads to more \emph{diverse} generated samples
since in each layer, properties are sampled from subspaces that ensure the disentanglement of variation along their basis. In other word, each dimension of a subspace capsule has unique effect on the generated samples. Figure (\ref{fig:mnist_tweak}) showcases this property of \emph{SCN}. Each row represents one feature like rotation, thickness of stroke, scale of generated digits and samples are generated by tweaking one dimension of capsules of the first layer  of generator in the range of $[-2.5,2.5]$. The generated samples in each row are diverse, and we can move over the appearance  manifold of each digit by changing the value of capsule dimension. Figure~\ref{fig:MNIST_Gen} shows the architecture of \textbf{SCN} generator with detailed training process explained in Section \ref{Exper}.  

\section {Projection Matrix Implementation} \label{Implementation_det}
The projection matrix $\boldsymbol{P}_c$ as defined in Equation~(\ref{Eq:projection_c}) involves taking the inverse of the square root of matrix $\boldsymbol{W}^T\boldsymbol{W}$, two very computationally expensive operations. If not being properly implemented, these operations can hinder the training process. In this work, we use an stable extension of Denman-Beavers iterative method~\cite{denman1976matrix}. 
It is known that for any symmetric positive (semi-)definite matrix $A$, there exists a unique symmetric positive (semi-)definite square root matrix.
Higham \etal proposed in \cite{higham1997stable} an iterative process that converges to the square root of such matrices. This iterative process is presented below:
Initialize $Y_0 = A$ and $Z_0 = I$. For $k=0,1,2,...$
 \begin{equation}\label{Schultz}
 \begin{aligned}
 Y_{k+1} = \frac{1}{2}Y_k(3I-Z_kY_K),\\
 Z_{k+1} = \frac{1}{2}(3I-Z_kY_K)Z_k,\\
 \end{aligned}
\end{equation}
where $k$ is iteration number. It has been shown that $Y_k$ and $Z_k$ converge to $A^{\frac{1}{2}}$ and $A^{-\frac{1}{2}}$, respectively. This process only requires matrix multiplication, which fits the best for parallel computation on GPUs. Further, it computes the inverse of square root of matrix $\boldsymbol{W}^T\boldsymbol{W}$ simultaneously. In all of our experiments we set the number of iterations as $k=20$. This iterative process increases the training time negligibly compared to the total training time. For instance, in our training of \emph{SCN} for large resolution images that all layers of generator are replaced by \emph{SCN} convolution layers, the training time is increased to 0.0529 sec/img compared to 0.047 sec/img for the baseline.

It is worth noting that when training process completes, the capsule projection matrix $\boldsymbol{P}_c$ is fixed and there is no time overhead for this iterative process.

\section{Experimental Results} \label{Exper}
In this section, we demonstrate the superiority of \emph{SCN}s on three tasks\footnote{Code:http://github.com/MarziEd/SubSpace-Capsule-Network}: Supervised classification of image data, semi-supervised classification of image data  and  generating high-quality images on multiple datasets.
\textbf{Datasets}: We use CIFAR10~\cite{krizhevsky2009learning}, Street View House Number (SVHN) \cite{netzer2011reading}, ImageNet \cite{imagenet_cvpr09}, CelebA \cite{liu2015faceattributes}, and 3 categories of  Lsun dataset, namely bedroom, cat and horse, throughout our experiments. 
\comment{
\begin{table}                                 
\small
\begin{center}
\small
\begin{tabular}{l||c|c}
\hline
\multirow{2}{*}{Methods} & CIFAR10 & SVHN \\\cline{2-3}
 & $N_l$= 4000 & $N_l$ =1000\\
\hline
\small {Improved GAN } \tiny{\cite{NIPS2016_6125}} & $18.63 ~\pm 2.32$ &$8.11~\pm 1.3$\\
\small {ALI } \tiny{\cite{dumoulin2016adversarially}}& $17.99~\pm 1.69$&$7.42~\pm 0.65$ \\ 
\small{LSAL} \tiny{\cite{edraki2018generalized}} & $16.22~\pm  0.31$&$5.46\pm 0.24$\\
\small{VAT} \tiny{\cite{miyato2018virtual}}& $14.87~\pm 0.13$&$6.83\pm 0.24$\\
\small{SCN} & $\mathbf{14.32}\pm\mathbf{0.21} $&$\mathbf{4.58}\pm \mathbf{0.18}$\\
\hline
\end{tabular}
\end{center}

\small{\caption{Classification errors on CIFAR-10 and SVHN datasets compared with the state-of-the-art methods. The error rates with $N_l$ = 4000 and $N_l$ = 1000 labeled training examples are reported.}}
\label{tbl:CIFAR10SVHN}
\end{table}
}
\subsection{SCNs for Classification}
\noindent\textbf{Semi-supervised classification}: For semi-supervised classification, we evaluate the performance of the  \emph{SCN} model on two benchmark datasets of CIFAR10 and SVHN through the GAN framework. To have a fair comparison with the state-of-the-art methods, we use the same network architecture and loss functions for generator and discriminator as the model proposed by~\cite{NIPS2016_6125} 

\textbf{SVHN}: In semi-supervised classification of the SVHN dataset, we replace the last 4 layers of the discriminator with subspace capsule layers. Figure \ref{fig:SVHN_disc} shows the architecture of \emph{SCN} discriminator. An input image passes through 6 convolutional layers that produce 128 feature maps of size $8\times8$. These feature maps go through three subspace capsule convolution layers, each layer has 64 different capsule types of 2-dimensional subspace. The first subspace capsule convolution layer has the kernel size of $3\times 3$ and the last two have kernel size of $1\times1$. We apply  the sparking function on all three layers. 
We feed the capsules of the last subspace capsule convolution layer to a subspace capsule mean pooling layer, with receptive field of $6\times 6$, that results in 64 capsule types of size 2, followed by the final subspace capsule fully connected layer with 10, 4-dimensional subspace capsule types. The input image belongs to the class with the largest norm of the output capsule.

\textbf{CIFAR10}: For the CIFAR10 dataset, the architecture of discriminator is similar to that of SVHN, except the subspace capsule convolution layers have 96 capsule types of size 2. 
The generator architecture for both datasets are the same as baseline architecture~\cite{NIPS2016_6125}.

We train the network using Adam optimizer with initial learning rate of $0.0003$ with $\beta_1=0.5$ and $\beta_2=0.99$. We hold out a set of 5000 training samples as our validation set for subspace capsule dimension selection, and fine tune the whole model on all training samples afterward.  
\begin{table}                                 
\small
\begin{center}
\small
\begin{tabular}{l||c|c}
\hline
\multirow{2}{*}{Methods} & CIFAR10 & SVHN \\\cline{2-3}
 & $N_l$= 4000 & $N_l$ =1000\\
\hline
\small {Improved GAN}\tiny{\cite{NIPS2016_6125}}& $18.63 ~\pm 2.32$ &$8.11~\pm 1.3$ \\
\small {ALI}\tiny{\cite{dumoulin2016adversarially} }& $17.99~\pm 1.69$&$7.42~\pm 0.65$  \\
\small{LSAL}\tiny{\cite{edraki2018generalized}} & $16.22~\pm  0.31$&$5.46\pm 0.24$\\
\small{VAT}\tiny{\cite{miyato2018virtual}}& $14.87~\pm 0.13$&$6.83\pm 0.24$\\
\small{SCN} & $\mathbf{14.32}\pm\mathbf{0.21} $&$\mathbf{4.58}\pm \mathbf{0.18}$\\
\hline
\end{tabular}
\end{center}
\caption{\small{Classification errors on CIFAR-10 and SVHN datasets compared with the state-of-the-art methods. The error rates with $N_l$ = 4000 and $N_l$ = 1000 labeled training examples are reported.}}
\label{tbl:CIFAR10SVHN}
\end{table}

Table \ref{tbl:CIFAR10SVHN} compares the performance of \emph{SCN} model on semi-supervised image classification of CIFAR10 and SVHN for 4000 and 1000 labeled samples, respectively.

\begin{table}                                      
\small
\begin{center}
\small
\begin{tabular}{|c|c|c|c|c|c|}
\hline
Model&Depth&SC-Fc&SC-Conv& Top1& Top5\\
\hline
Resent&34&-&-&27.13&8.84 \\
SCN &34&(1000,4)&~(256,2)&\textbf{25.64}&\textbf{8.17} \\
SCN &34&(1000,4)&~(128,4)&25.96& 8.35\\
\hline
\end{tabular}
\end{center}
\caption{\small{Single crop, Top1 and Top 5 error rate of ImageNet classification with Resnet backbone model. 
In SC-Fc and SC-Conv columns, in a tuple $(n,c)$ , $n$ is the number of capsule types and $c$ is the subspace capsule dimension.}}
\label{tbl:ImageNet }
\end{table}
 \begin{figure*}
 \begin{center}
\subfigure[CelebA]{

\begin{minipage}{.8\textwidth}
  \centering
  \includegraphics[width=0.8\linewidth]{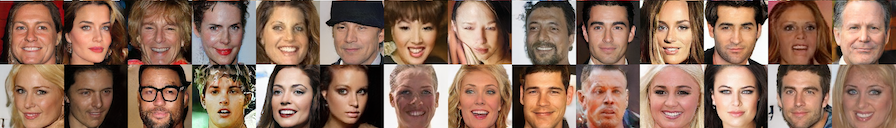}
  \label{fig:celebA}
\end{minipage}}
\subfigure[LSUN-bedroom]{
\begin{minipage}{.8\textwidth}
  \centering
  \includegraphics[width=0.8\linewidth]{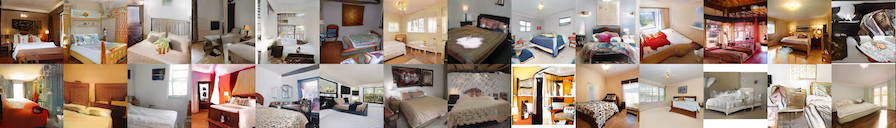}
  \label{fig:bedroom}
\end{minipage}}
\subfigure[LSUN-horse]{
\begin{minipage}{.8\textwidth}
 \centering
  \includegraphics[width=0.8\linewidth]{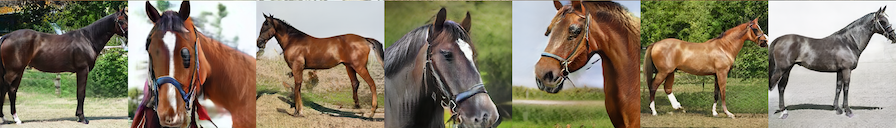}
  \label{fig:horse}
\end{minipage}}
\subfigure[LSUN-cat]{
\begin{minipage}{.8\textwidth}
  \centering
  \includegraphics[width=0.8\linewidth]{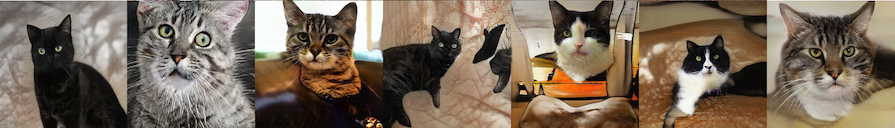}
  \label{fig:horse}
\end{minipage}}
\end{center}
\caption{Generated samples for various datasets. (a) CelebA $(128\times 128)$, (b) Bedroom $(128\times 128)$, (c) Horse $(256\times 256)$. (d) Cat $(256\times 256)$.} 
\label{fig:celeba_bedroom_horse}

\end{figure*}

 \noindent\textbf{Supervised classification}: We evaluate the scalability of \emph{SCN} on large datasets like ImageNet. We also compare the performance of \emph{SCN} with capsule network proposed by Sabour \etal. on CIFAR10 dataset.
 
 \textbf{ImageNet:} For ImageNet dataset, the last 4 layers of the Resnet model with depth of 34 have been replaced with SCN layers, batch normalization layers in the final block and also the final residual connection are removed. Mean pooling is replaced by the SCN mean pooling. The model was trained using SGD with momentum rate of 0.9 for 100 epochs. The learning rate is initialized as 0.1 and decayed every 30 epochs with the rate of 0.1. Table \ref{tbl:ImageNet } shows that \emph{SCN} outperforms the baseline model and reduces the relative top-1 classification error of Resnet by $5\%$. 
 
 \textbf{CIFAR10:} For supervised classification of CIFAR10, we also update the convolution layers of the last bottleneck block of Resnet model with 110 layers to SCN convolution layers. Each of them has 32 capsule types with subspace capsule dimension $c=2$. Batch normalization layers and residual connection of this block has been removed. Mean pooling is replaced by SCN mean pooling and the final fully connected layer is replaced by SCN fully connected layer with 10 capsule types with subspace capsule dimension $c=4$. This model archives $5.15\%$ error rate that significantly outperforms capsule network model \cite{sabour2017dynamic} with $10.6\%$ error rate. It also improved the relative error rate of the Resent model by $19.6\%$ by reducing it from $6.41\%$ to $5.15\%$  
\subsection{SCNs for Image Generation}
We evaluate the effectiveness of subspace capsule networks on the image generation task using the GAN framework for various size of images and datasets. In all of our experiments, we build the generator based on subspace capsule networks and the discriminator based on CNNs.
 \begin{table}                                     
\small
\begin{center}
\small
\begin{tabular}{|c|c|c|c|c|}
\hline
Method&CelebA&Bedroom& Horse&Cat\\
\hline
ProgGAN\tiny{\cite{karras2017progressive}}&$9.67^*$&$21.1^*$&$16.11$&$37.52$\\
\hline
SCN& $\textbf{6.23}$&$\textbf{9.94}$&$\textbf{12.83}$&$\textbf{29.20}$ \\

\hline
\end{tabular}
\end{center}
\caption{\small{Comparison of FID score of \emph{SCN}
with our baseline model. Entries with $*$ are our rerun of the baseline.}}
\label{tbl:FIDscore}
\end{table}

\textbf{MNIST}: The \emph{SCN} architecture of generator is shown in Figure \ref{fig:MNIST_Gen}. The first layer has 10 subspace capsule types. Each of them is a $16-$dimesional capsule subspace. The output of the first layer is 10 subspace capsules. The capsule with the largest output vector is selected and reshaped to a $(2 \times 2\times 25)$ tensor. 
This tensor goes through a bilinearly upsampling layer to double the spatial size and a subspace capsule convolution layer with kernel size of $(8,16,3,3)$. The third layer has
the same structure of upsampleing and subspace capsule convolution layer as the second layer except that it has the kernel size of $(8,8,3,3)$. This is followed by the last subspace capsule convolution layer with kernel size of $(8,8,3,3)$. The final layer is a transposed convolution layer with the receptive field of $(5\times 5)$ with stride of 2 followed by sigmoid activation function. All subspace capsule convolution layers have stride 1 and squashing activation function. 
The discriminator architecture is composed of 4 convolution layers with receptive field of $(5\times 5)$ and stride of 2. We apply batch normalization to all convolutional layers and the activation function is leaky Relu with slope of $0.2$. This is followed by a global mean pooling and a fully connected layer to 10 output classes.

We follow AC-GAN \cite{odena2017conditional}, and add an auxiliary classification loss to ensure that each capsule subspace in the first layer of generator captures the variation of a single class. To this end, we use the index of the capsule with the maximum length in the first layer as the ground truth label for the generated sample.
We train this model using Adam optimizer with initial learning rate of 0.0002 for 25 epochs. 

\textbf{High-Resolution Images}: We also apply \emph{SCN} for generating high-resolution images of size $128^2$ and $256^2$ for CelebA and 3 classes of LSUN datasets. To have a fair comparison with the state-of-the-art models, we build \emph{SCN} generative model based on the model proposed by \cite{karras2017progressive}. Karras \etal suggest to use progressive growing of generator and discriminator models for generating high resolution images. The training starts from a low resolution of $4\times4$ and gradually a new block for higher resolution is added to the both generator and discriminator models. This process continues until the networks get to the final resolution of images. Each block in this model consists of an up-sampling step and a convolutionl layer. 

 For CelebA and LSUN bedroom datasets we generate samples with resolution of $128^2$. In the generator model, we update all the convolutional layers from resolution $4$ to resolution $64$ to \emph{SCN} convolution, and Relu activation function is replaced by squashing activation function. The higher resolution blocks of $64$ and $128$ are remained intact. 
 For LSUN cat and LSUN horse datasets, we generate samples of size $256^2$. In the generator network, we replace all convolutional layers for all resolutions with \emph{SCN} convolution layers followed by squashing activation function. Table \ref{tbl:Capsule_dim} presents the configuration of subspace convolutional layers for all experiments. We use the tuple notation of $(n,c,k,k)$ to denote a subspace capsule convolution layer with $n$ capsule types, $c$-dimensional capsule subspaces and a receptive field of $k\times k$.
 
 To stabilize training process we adopt Wasserestien loss with gradient penalty. We also benefit from progressive growing through training process. 
 For all of the experiments, the discriminator network is the same as the baseline architecture.
 \begin{table}                                            
\begin{small}
\begin{center}
\small
\begin{tabular}{|l||c|c|}
\hline
\multirow{2}{*}{LR} & CelebA, bedroom &cat, horse \\\cline{2-3}
 &FR=128& FR=256\\
\hline
4&~(4, 128, 3, 3)& (8, 64, 3, 3)\\
8&~(4, 128, 3, 3)&(8, 64, 3, 3) \\
16&~(4, 64, 3, 3)&(8, 64, 3, 3)\\
32&~(2, 64, 3, 3)&(8, 64, 3, 3)\\
64&-&(8, 32, 3, 3)\\
128 &-&(4, 32, 3, 3)\\
256 &-&(2, 32, 3, 3) \\
\hline
\end{tabular}
\end{center}
\caption{Configuration of subspace capsule convolution layers for the generator networks. ``LR" and ``FR" stand for the layer resolution and  final image resolution respectively.}
\label{tbl:Capsule_dim}
\end{small}
\end{table}
\begin{figure*}
\begin{center}
\includegraphics[width=0.70\linewidth]{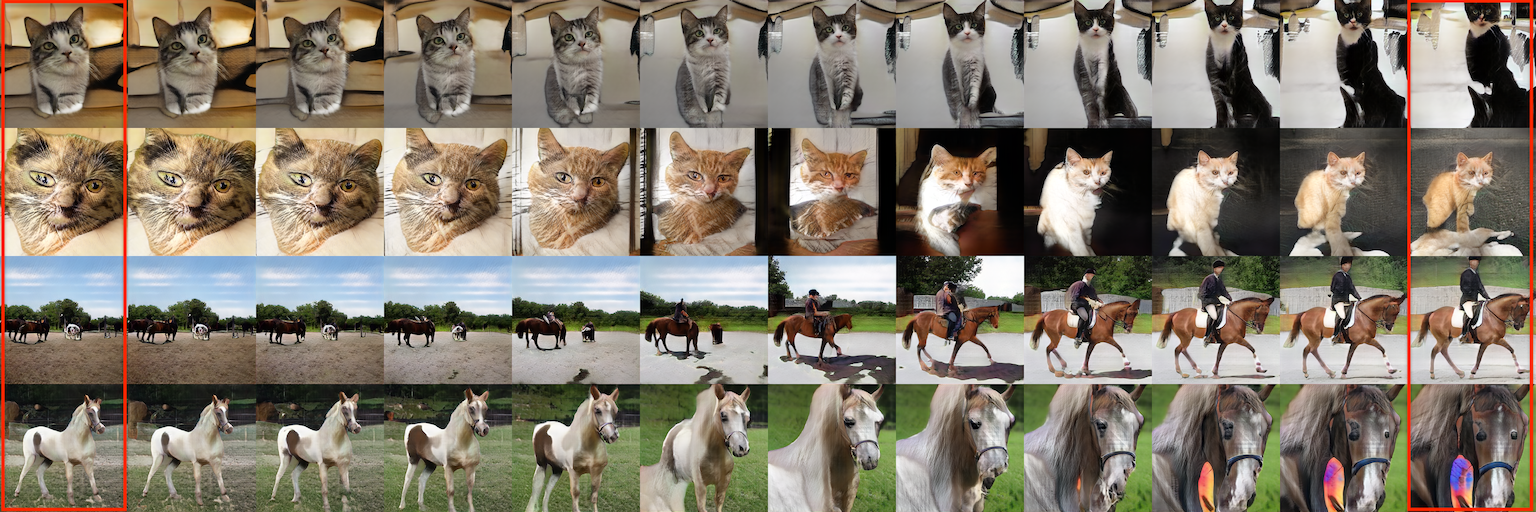}
\end{center}
   \caption{ In each row, the first and last samples in the red boxes are generated using two independent noise vectors. The intermediate samples are generated by walking through the linear interpolant of those two noise vectors. }
\label{fig:Interpolate}
\end{figure*}
We compare the generated samples quantitatively with the state-of-the-art model using Fréchet Inception Distance (FID). We believe the FID metric is the closest one to the human judgment, since it compares the statistic of features extracted; using Inception model; from generated samples with real samples. Comparison of numerical values of this metric for all datasets are presented in Table~\ref{tbl:FIDscore}. In all 4 datasets, \emph{SCN} consistently improved the relative FID score of generated samples by at least $20\%$. Figure (\ref{fig:celeba_bedroom_horse}) shows generated samples for these datasets.

\textbf{Interpolation of Latent Representation}: To verify that \emph{SCN} generator does not merely memorize training samples, we also walk through the manifold space. To this end we choose two random latent representation ${z}_1$ and ${z}_2$, then we use SCN generator to generate samples for ${z}$ s on the linear interpolant of ${z}_1$ and ${z}_2$.
Figure (\ref{fig:Interpolate}) shows the interpolated samples for LSUN-horse and LSUN-cat datasets. As it can be seen the transition between pairs of latent representations are smooth and meaningful.
 \subsection{Ablation Study}\label{Ablation}
 In this section we analyze the effect of subspace capsule size and also position in the network on performance. 
 
 Table 5 reports the semi-supervised classification error rate of SVHN dataset with 1000 labeled training samples, when we update the last fully-connected or convolution layers with various size capsules. Configuration 0 demonstrates the result of the baseline model \cite{NIPS2016_6125},
 the first three rows after that correspond to the settings when subspace capsules are only applied on the last layer with various capsule sizes of 2,4 and 8. The configurations [4-6] correspond to the settings when the last 3 convolution layers are replaced in the discriminator with subspace capsule convolution layers. We conclude the following results from this analysis. 1- Subspace capsule layers are effective in improving the overall performance even if we use them only in one layer of the discriminator network. 2- The proper combination of capsule types and the capsule dimension plays a key role in achieving the best performance.
  \section{Conclusion}\label{conclud}
In this paper, we proposed SubSpace Capsule Networks, referred to as \emph{SCN}s, which offer a general capsule model with no computational overhead compared to CNNs. \emph{SCN} learns a group of capsule subspaces to model the variations in the properties of an entity through the sequence of layers.
We successfully applied \emph{SCN} on the GAN framework, both on generator and discriminator networks leading to the state-of-the-art performance in semi-supervised classification on CIFAR10 and SVHN and significantly improving the quality of generated samples.  
 \begin{table}                                            
\small
\begin{center}
\small
\begin{tabular}{|c|c|c|c|}
\hline
config&SC-Fc&SC-Conv& Error rate\\
\hline
0 \tiny{\cite{NIPS2016_6125}}&-&~-&8.11 \\
\hline
1&(10,2)&~-&5.8 \\
2&(10,4)&~-&5.12\\
3&(10,8)&~-&5.2\\
4&-&(64,2)&5.26\\
5&- &(32,4)&5.49\\
6&- &(16,8)& 5.37 \\
7&(10,4)&(64,2)& 4.58\\
\hline
\end{tabular}
\end{center}
\caption{Error rate of semi-supervised classification for SVHN dataset for 1000 labeled samples for various size and type of subspace capsule. SC-Fc stands for subspace capsule fully connected layer. In a tuple (n,c), n is the number of capsule types and c is the subspace capsule dimension.}
\label{tbl:abalation-semi-supervised}
\end{table}
\section{Acknowledgments}
This research is based upon work supported in parts by the National Science Foundation under Grants No. 1741431 and Office of the Director of National Intelligence (ODNI), Intelligence Advanced Research Projects Activity (IARPA), via IARPA R$\&$D Contract No. D17PC00345. The views, findings, opinions, and conclusions or recommendations contained herein are those of the authors and should not be interpreted as necessarily representing the official policies or endorsements, either expressed or implied, of the NSF, ODNI, IARPA, or the U.S. Government. The U.S. Government is authorized to reproduce and distribute reprints for Governmental purposes notwithstanding any copyright annotation thereon.
We also would like to thank Dr. Jun Wang for generously  providing us access to the CASS GPU cluster supported in parts by the US Army/DURIP program W911NF-17-1-0208.
\begin{small}
\bibliographystyle{aaai}
\bibliography{egbib}
\end{small}
\end{document}